\documentclass[a4paper, 10pt]{report}
\usepackage[bindingoffset=0.2in,left=0.55in,right=0.55in,top=1in,bottom=1in,footskip=.25in]{geometry}
\usepackage{bm}
\usepackage{amsfonts}
\usepackage{subfig}
\usepackage{amsmath, amstext}
\usepackage{amssymb}
\usepackage{graphicx}

\begin{document}

\title{\Huge Multi-Sensor Event Detection using Shape Histograms}
\author{\LARGE Ehtesham Hassan \\ ehtesham.hassan@tcs.com
\and
\LARGE Gautam Shroff \\ gautam.shroff@tcs.com
\and 
\LARGE Puneet Agarwal \\ puneet.a@tcs.com
\and \\
\LARGE TCS Research, New Delhi, India}

\date{\today}

\maketitle
\begin{abstract}
Vehicular sensor data consists of multiple time-series arising from a number of sensors. Using such multi-sensor data we would like to detect occurrences of specific \textit{events} that vehicles encounter, e.g., corresponding to particular maneuvers that a vehicle makes or conditions that it encounters. 
Events are characterized by similar waveform patterns re-appearing within one or more sensors. Further such patterns can be of variable duration. In this work, we propose a
method for detecting such events in time-series data using a novel feature descriptor motivated by similar ideas in image processing. We define the \textit{shape histogram}: a constant dimension descriptor that nevertheless captures patterns of \textit{variable} duration. We demonstrate the efficacy of using shape histograms as features to detect events in an SVM-based, multi-sensor, supervised learning scenario, 
i.e., multiple time-series are used to detect an event. We present results on real-life vehicular sensor data and show that our technique performs better than available pattern detection implementations on our data, 
and that it can also be used to combine features from multiple sensors resulting in better accuracy than using any single sensor.
Since previous work on pattern detection in time-series has been in the single series context, we also present results using our technique
on multiple standard time-series datasets and show that it is the most versatile in terms of how it ranks compared to other published results.
\end{abstract}

\section{Introduction}
\label{sec:intro}
While engineering new vehicles quality engineers schedule trial runs in a controlled environment to simulate the
life cycle of a vehicle in the field. In order to do this effectively, it is important to understand how similar models are used in the field by customers,
especially in terms of the distribution of certain maneuvers that stress various components, such as rapid acceleration,
sudden braking, sharp turns, etc. 
The problem reduces to detecting occurrences of specific events from time-series data
obtained from multiple on-board sensors across numerous runs of a large population of vehicles.
 
Training data was generated by running test vehicles in a controlled environment where the maneuvers of interest are
performed and their time-stamps are logged.
The task is to detect similar events in large collections of time-series data coming from vehicles in the field, efficiently and with high accuracy. 
In current practice engineers use hand-crafted rules to detect such events, for example ``if sensor-1 is higher than $t_1$ and derivative of sensor-2 is
more than $t_2$'', occurrence of an event is assumed to be indicated. However, it was found that 
such rules could not discover many occurrences of actual events: for example, we observed
only 60-70\% accuracy when using the rule based approach.

Pattern recognition in time series, even in the single sensor case, is a well-studied research problem. The problem at hand is 
is further complicated because of the following practical challenges as observed in real-life data: a)~the duration of an event is not fixed, b)~the magnitude of the
events may vary, c)~sensor readings are often sparse, d)~if the event is divided into sub-sections, the ratio of the durations of
these sub-sections varies in different instances of the same event, and e)~these events need to be detected from large time-series,
therefore linear time algorithm is required. It is because of such challenges that the available techniques were found not to perform as well
as needed, in terms of accuracy, efficiency or both.

The problem can be viewed as that of supervised classification in interesting and non-interesting categories since
known occurrences of the specific patterns are available.  However, previous work on pattern detection in time-series
\cite{Guralnik:1999,Hunter99knowledge,Wei:2006,Ye:2009} have primarily been on single time-series and also do not address 
some of the above challenges. In this paper we present a method for detection of time-series events that addresses these challenges. 

We introduce a novel descriptor, the \textit{shape histogram}, for representing of variable length time-series subsequences by
exploiting the \textit{shape} of a temporal pattern. The concept is borrowed from similar ideas in image processing, applied here in the
one-dimensional world of time-series analysis. The shape histogram generates a time-scale invariant representation of time-series subsequences
and is conveniently usable with various distance measures and learning algorithms.  We have applied our proposed shape histogram 
to the specific task of event detection from time-series by using it as a feature vector for supervised classification. 

Modeling time-series events requires exploiting long range correlations together with
the short range interactions within subsequences. Our shape histogram descriptor addresses this problem by defining a shape-based
representation of patterns that represents variable duration subsequences by a two-dimensional vector. 
The shape-histogram is a robust representation of time-series subsequences tolerant
to variable time-scales as well as missing values. 

Shape-histogram-based event detection is independent of expansions, compressions and discontinuities due to the inherent non-linearity embedded
in the subsequence generation process in comparison with existing works which are applicable to constant duration time-series patterns. Also, the
feature vector derived from a shape histogram can be easily combined with other features such as the duration or amplitude of an event. 
Thus we we were able to use multiple sensors for our classification task, which results in improvements over any single sensor approach.
We also demonstrate that our shape histogram based approach is most versatile and has the best rank as compared to various approaches available in
research literature, on multiple standard time-series datasets. 

The time complexity of generating a shape histogram from a time-series
subsequence is quadratic in the size (duration) of the subsequence.
However, subsequences, albeit of variable length, are always much smaller than the length of the time series and can be 
assumed to be bounded by a constant.
Computing the shape histogram for all subsequences in a time-series is thus a linear-time procedure.
Since the size of the shape-histogram is fixed and small, a discriminative classifier, such as SVM,
also works efficiently. 

Lastly, our method of feature extraction is un-supervised and performs better in both
efficiency and quality even as compared to supervised methods of feature extraction such as shapelets\cite{Ye:2009,Mueen:2011,Lines:2012}
that exploit knowledge of class labels while deriving features, which we do not assume.

The document is structured as follows: We begin in Section \ref{sec:prob} with an overview of our definition of multi-sensor time-series events
and challenges faced in practice with vehicular sensor data. 
We motivate and introduce our novel shape-histogram feature descriptor in Section \ref{sec:sh}. 
Experimental evaluation of our technique on a real-life vehicle sensor data as well as standard datasets is presented in Section
\ref{sec:exp}. Section \ref{sec:rw} describes related work in time-series pattern detection.

\section{Overview}
\label{sec:prob}
Previous results in time series mining can be broadly categorized in two groups. 
The first is to discover patterns (motifs) in an unsupervised manner, often using clustering of time-series
subsequences\cite{Rakthanmanon:2011}. 
The second set of approaches exploit domain knowledge and available training data for supervised classification of subsequences in order to
detect specific events characterized by particular waveform patterns\cite{Lines:2012}. We are concerned with the latter problem.

Here we formally define a time-series, and a subsequence. A time series $t_i$ is an ordered set of $m$ values, i.e., $t_i$ = \{$q_1$, ...,
$q_m$\}. The sequential order of values characterizes their temporal ordering with respect to a predefined sampling rule. 
Often there are multiple sensors in a system and more than one of the sensors may characterize the target event. Therefore we assume that
there are multiple such time-series corresponding to different sensors, i.e., $T_i = \{\tau^1_i, ..., \tau^j_i, ..., \tau^s_i\}$ for $s$ sensors.
We get such a time-series from every session of operation of the underlying system, i.e., one set of time-series $T_i$ for every independent
run of a vehicle. Assuming there are $n$ such runs / time-series, we have time-series dataset $T = \{T_1, ..., T_i, ..., T_N\}$. For single
sensor time-series we use a shorthand $\tau_{i}^j$, which refers to a time-series from $i^{th}$ run of a vehicle and $j^{th}$ sensor.

\begin{figure}
\centering
\includegraphics[width=0.6\columnwidth]{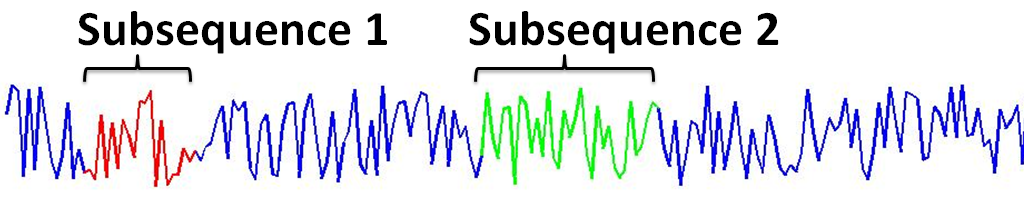}
\caption{An example time series generated with random values and two example subsequences}
\label{fig111}
\end{figure}

A subsequence $S$ is a smaller section of $\tau_{i}^{j}$. Formally, a subsequence $S$ consists of
a set of \textit{approximately} $l$ contiguous positions from $\tau_{i}^{j}$, i.e., $S$ = \{ $q_i$, $q_{i+1}$, ..., $q_{i+l-1}$\}. 
We also assume that length of such subsequences is not fixed, i.e., the length of subsequences $l = |S|$
can vary, albeit being bounded by a maximum value considerably smaller than the length of each time series.
This definition captures the general case where no condition is imposed on the
length of each subsequence or the overlap between two subsequences (refer Figure \ref{fig111}). In this setting we define the problem of event
detection in time-series as follows:

Given a training set of (potentially variable length) subsequences labeled as interesting or non-interesting , i.e., $\{1,0\}$, the objective is to 
classify a set of unlabeled candidate subsequences (also of variable length) that have been extracted from a set of time-series $T$. 

\subsection{Solution Overview}
\label{sec:soln}
The first step of the solution is related to identification of candidate subsequences from the time-series dataset $T$. Most previous works
have used constant window size subsequences, which requires some way to choose the correct or optimum window size within which all
occurrences of the desired patterns will fall. However, in practice, the patterns characterizing an interesting event are present in
subsequences of different length that are all nevertheless visually similar. 

For defining the boundary of a subsequence we assume that hand-crafted rules based on domain-specific heuristics are available,
which is indeed the case in practice.
We illustrate the use of such rules in the case of time-series obtained from multiple sensors available in a commercial vehicle
operating in varying operating conditions. Details of the dataset is described in Section \ref{sec:exp}, and our method of identifying
candidate subsequences, and an example of vehicular event \textit{Event1} is given in Appendix-A.

\begin{figure}[!h]
\centering
\includegraphics[width=0.82\columnwidth]{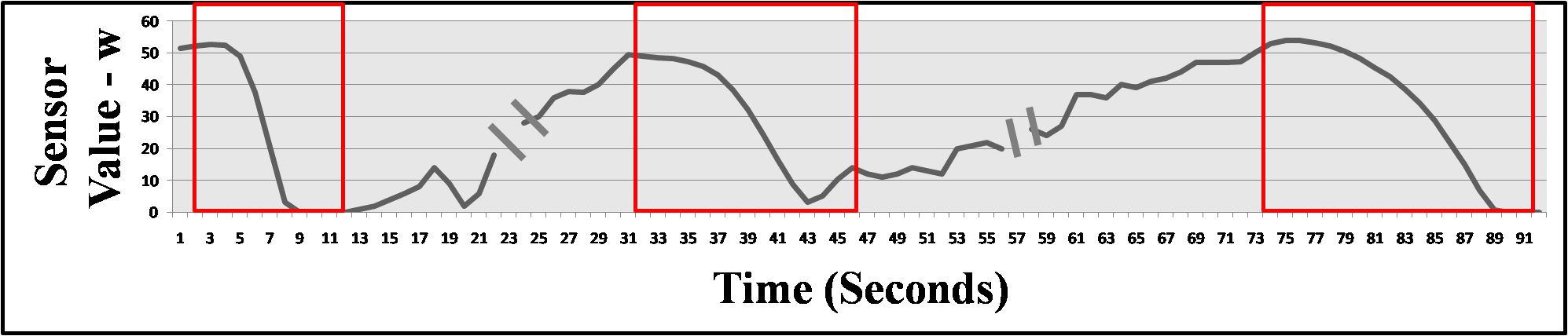}
\caption{Potential occurrences of \textit{Event1} in \textit{Sensor2}.}
\label{Ex_events}
\end{figure}

\begin{figure}
\centering
\includegraphics[width=0.85\columnwidth]{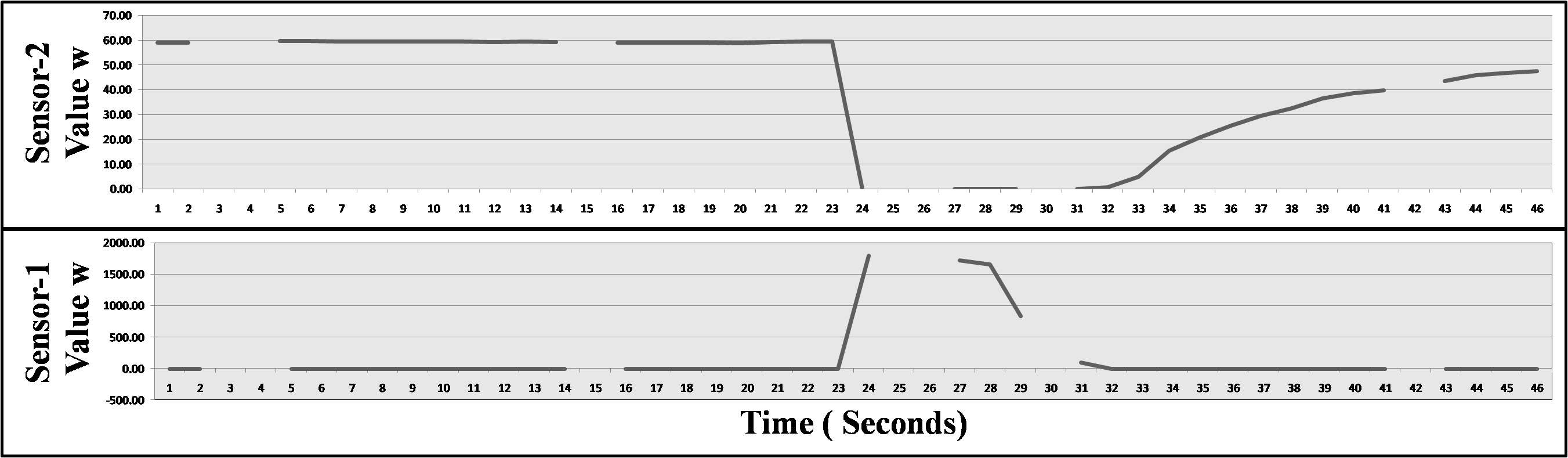}
\caption{Sensor reading (y-axis) for \textit{Event1}, time on x-axis.}
\label{fig_hard_stop_event}
\end{figure}

Figure \ref{Ex_events} depicts multiple occurrences of the same event, but of variable time duration.
Further, in many practical scenarios, real-time measurements have missing values because sensors sometimes
fail to capture data, especially for extreme values, as can be observed in Figure \ref{fig_hard_stop_event} taken from our vehicle
sensor data. Such regions of missing values also occur multiple times and are of varying duration within an event.

After extraction of candidate subsequences we construct a feature vector for every subsequence using the
shape-histograms (described in Section \ref{sec:sh}). A key property of the feature vector drawn from shape
histogram is that it can be combined with shape histogram of other sensors. For this we collate subsequences from other sensors sharing
same time-stamp and concatenate their shape-histogram features vectors, as well as add additional features if desired.
Finally we train an SVM classifier using the training data and use it to classify the candidate subsequences into interesting and uninteresting categories.
\section{Shape histogram}
\label{sec:sh}
\subsection{Motivation} 
Developments in image processing and computer vision have produced a variety of shape descriptors that exploit distinct attributes such as
shape, appearance, and texture to develop feature-based object representation. Some of the well established descriptors are discussed in
\cite{ICPR10_R4,DAS_R22,MM:Lowe,ICCVWSRef_29,ICCVWSRef_30,SD_Hassan} have been shown to result in robust object representations that display
low intra-class and high inter-class variance in different application scenarios. 

\begin{figure}[!h]
\centering
\includegraphics[width=0.65\columnwidth]{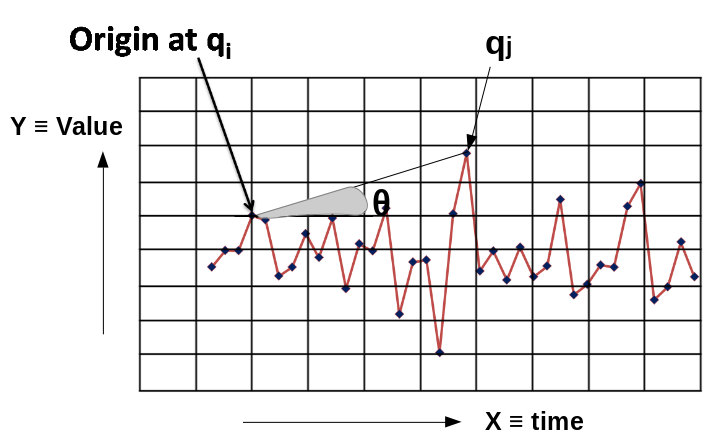}
\caption{Time-series subsequence $S$ plotted in $\mathbf{R}^2$}
\label{fig1}
\end{figure}

In time-series, however, patterns are uniquely characterized by the sequential order of numerical or symbolic attributes, where
the underlying temporal order between consecutive values defines a highly correlated structure. Each point $q_i$ of a
time-series subsequence $S$ is defined by two attributes: 1) its value ($w$), and 2) its temporal order ($t$), i.e., $q_i = (w_i, t_i)$. Figure
\ref{fig1} shows an example subsequence in which two points $q_i$ and $q_j$ are highlighted. The value $w$, and temporal
order $t$ of a point $q_i$ can be treated as coordinates of a point  in two-dimensional Euclidean space, in which the subsequence
defines a curve. The intuition is that curves of similar shape in this two-dimensional space should correspond to subsequences 
representing the same event. Motivated by the success of shape based object descriptions in image processing, such as the
`shape context'\footnote{\label{wikishape}http://en.wikipedia.org/wiki/Shape\_context}\cite{ICPR10_R4}, 
we define a 
similar representation for time-series patterns that we call the \textit{shape histogram}.

\subsection{Definition}
We assume the prior availability of subsequences filtered from a large time-series via rule based techniques as described in Section
\ref{sec:soln} and Appendix-A. 

The shape histogram ($H$) for a subsequence $S$ captures the distribution of all pairs of points (i.e., in two dimensional value, time space)
with respect to their relative distances and angles. For example, in Figure \ref{fig1}, the relative angle of $q_j$ from $q_i$ is $\theta$,
computed with respect to the X-axis with origin at $q_i$. Similarly
the distance of $q_j$ from $q_i$ is the euclidian distance in two dimensional value-time space. We compute 
the angle and distance between each pair $(q_i,q_j), q_j \neq q_i$. \textit{Note:} One may think that it is sufficient to use either $w$
or $t$ instead of using both in euclidean distance, because the other dimension is based on angular distance. However, after a careful
observation of extrema cases $(w_i-w_j=0 \text{ OR } t_i-t_j=0)$ it can be understood that it is important to use both in the distance
dimension.

These relative pairwise angles and distances encode the trend in sensor values with respect to time and form a
two-dimensional distribution which is the shape histogram. Given a subsequence $S= \{q_1, ..., q_{i}, ..., q_l\}$ we define
its multi-dimensional \textit{shape histogram} $H \in \mathbb{N}^{m \times n}$  as follows:
\begin{equation}
H(\mathbf{k}) = [\#\{(q_i,q_j) \in bin(\mathbf{k})\} | (q_i \neq q_{j})]
\end{equation}

Here $\mathbf{k} \in \mathbb{N}^{2}$ and 
$H(\mathbf{k})$ is the $\mathbf{k}^{th}$ values of the shape histogram $H$, and it is the number of pairs $(q_i,q_j)$
that have the relative angular and euclidean distances within a given range, i.e., that fall in $bin(\mathbf{k}$,
which defines the distance and angle ranges of $\mathbf{k}^{th}$ $bin$ of the histogram. 
We shall describe how the ranges for each bin are computed in Section \ref{sec:comp}.

Note that some shape descriptors, also based on shape context, compute the distances in log-polar space \cite{ICPR10_R4, SD_Hassan}. 
However, we want the shape
histogram for a time-series subsequence to be equally sensitive to near and distantly placed points, as the dominance of local versus global shape
properties varies from pattern to pattern.
Therefore, the proposed shape histogram is based directly on the distance and angle rather than log-polar coordinates. 
We also assume the points $(q_i)$ are uniformly spaced on the time-scale. 

\subsection{Computation}
\label{sec:comp}
The shape histogram in the context of time-series is a two dimensional histogram, with the dimensions being along the distance
bins, and angle bins. We assume it has $m$ distance bins and $n$ angle bins. Sample shape histograms are shown in Figure
\ref{fig31}. 

For each point $q_i \in S$, we compute its distance and angle with the remaining points of $S$ in two-dimensional Euclidean space. 
In the angle computation for $q_i$, origin is set at $q_i$, and orientation of remaining points is computed with respect to the X axis. Figure
\ref{fig1} shows point $q_j$ located at the angle $\theta_{ij}$ from $q_i$, computed with origin at $q_i$. The distance between points
$q_i \equiv (w_i, t_i)$ and $q_j \equiv (w_j, t_j)$ is defined as $d_{ij} = \sqrt{(w_i-w_j)^2 + (t_i-t_j)^2}$. 

Repeating the procedure for all points, we get the distance matrix $D = [d_{ij}]$ and angle matrix $A=[\theta_{ij}]$ where each row corresponds to a point $q_i$ in
$S$,  We divide each row of the distance matrix by maximum distance in that row to normalize it to a $[0, 1]$ scale. The angles are anyway
defined on a fixed scale of 0 to 359.99 degrees so they do not require normalization. The shape histogram $H$ for a subsequence $S$ will be a
two-dimensional histogram representing the joint distribution of elements in $D$ and $A$, computed as follows.

A shape histogram calculated for all pairs of points, keeping one of the points $q_i$ fixed, also has same size $(m\times n)$ and is
referred as the \textit{shape context} of the point $q_i$. First, we compute the shape context $H_i$ of each point $q_i$ defined by its
distance-angle based distribution with respect to the remaining points. Considering $[d_0, d_1]\cup[d_1, d_2]\dots\cup[d_{m-1}, d_m]$ as the
bins for distance quantization, and $[\alpha_0, \alpha_1]\cup[\alpha_1, \alpha_2]\dots\cup[\alpha_{n-1}, \alpha_n]$ as the bins for angle
quantization, i.e., quantizing the distances in $m$ levels and angles in $n$ levels. 
(Note that all distance bins, and angle bins are of uniform size.) For point $q_i$, the count of the $bin(x, y)$ of its shape context is
defined as 
\begin{equation}
H_i(x, y) = \sum_{j = 1, j \neq i}^{l}\phi(D_{i,j}, A_{i,j}, d_{x-1}, d_x, \alpha_{y-1}, \alpha_y), 
\end{equation}
where
\begin{eqnarray}
\phi(D_{ij}, A_{i,j}, d_{x-1}, d_x, \alpha_{y-1}, \alpha_y) \hspace{30mm}\nonumber \\ \hspace{10mm} = \left\{
\begin{array}{l l}
1 & \quad \textup{if} \quad D_{i,j}\in[d_{x-1}, d_x],\ A_{i,j}\in[\alpha_{y-1}, \alpha_y] \\
0 & \quad \textup{otherwise}
\end{array}\right. \nonumber
\end{eqnarray}

Summing the shape contexts of all points in $S$ i.e., $H_{i}(:,:)$ $\forall i \in {1, ..., l}$ defines the global distribution of the
point-pairs. The summation $H_{sum} = \sum_{i}H_i$ represents distribution of euclidean distances and angles between the points
in $S$. In $H_{sum}$, count of each $bin(x,y)$ represents the count of pair points, which are arranged within distance $[d_{x-1},
d_{x}]$ and angle $[\alpha_{y-1}, \alpha_y]$. Dividing $H_{sum}$ by its maximum entry we arrive at the shape histogram $H$ for subsequence
$S$. 

Next, the value of every bin of $H$ is then identified as a feature for discriminative classification, therefore we convert the
shape histogram $H$ into $p=m\times n$ dimensional feature vector. 

Further, a very important property of the shape histograms is that they can be combined with other features extracted from the same
time-series, such as average value $\overline{w_i}$ of the subsequence. Not just this, additional features can be drawn from the shape
histograms of other sensors of the same system, and can be combined in order to improve the accuracy of time-series event
detection. This property facilitates the usage of a discriminative classifier and we call this method of combining the shape histograms
with other features or other shape histograms as \textit{linear concatenation}. We demonstrate this aspect of shape histograms through
experiments on a real-life dataset in Section \ref{sec:e2}.

\begin{figure}[!h]
\centering
\includegraphics[width=0.85\columnwidth]{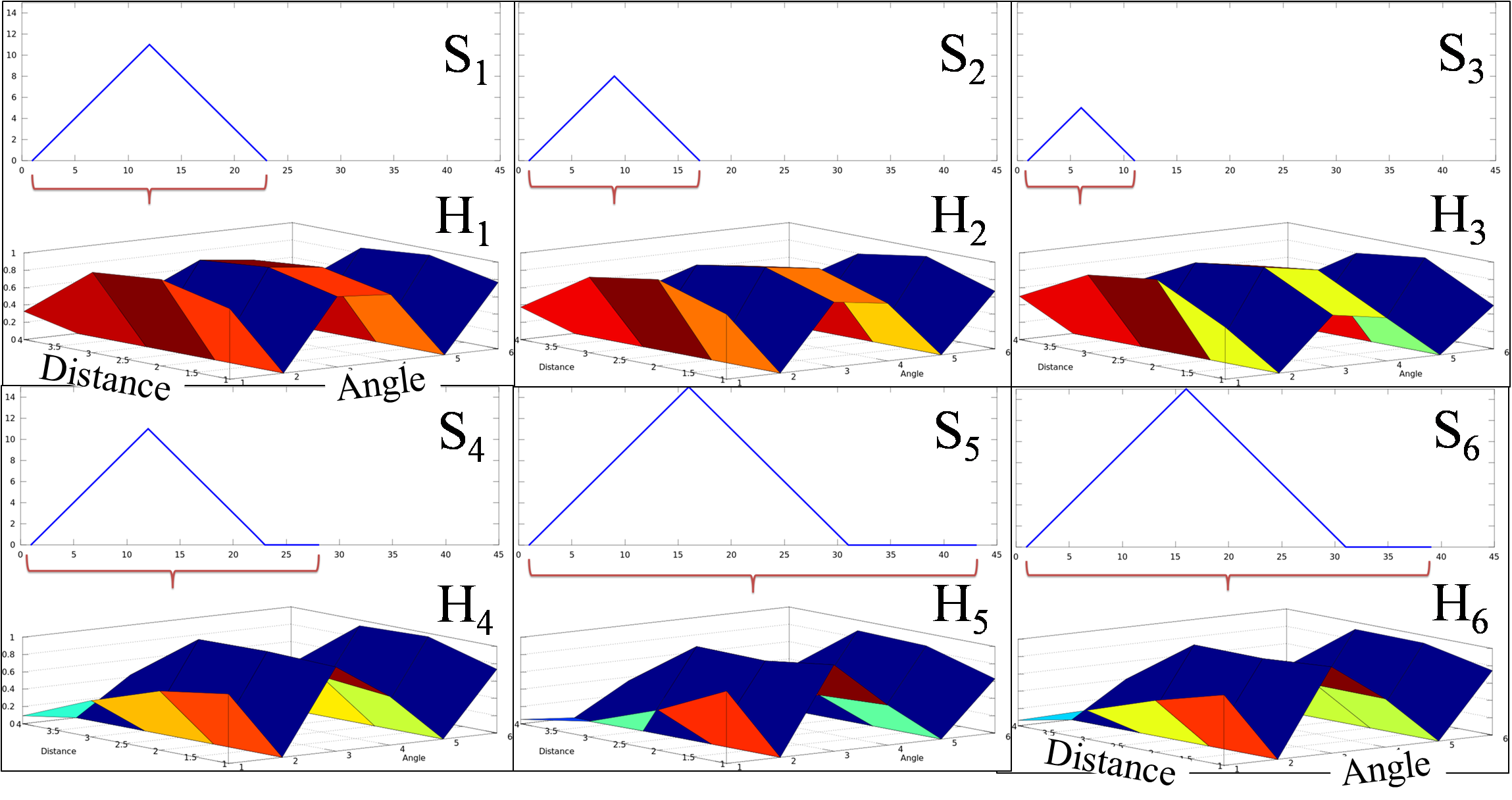}
\caption{Contrived example of subsequences ($S_1$ - $S_6$) and surface plots of corresponding shape histogram(($H_1$ - $H_6$)) placed below.
Each subsequence is of different length limited in the underneath brace. (Axis values can be ignored.)}
\label{fig11}
\end{figure}

\subsection{Examples and Intuition}
Notice that $H$ is invariant to $l = (|S|)$ making it applicable for time-series subsequences of variable length. Figure \ref{fig11}
shows shape histogram for some example subsequences to establish their scale invariance. 
In other words, the point-pair distribution histogram encodes the temporal arrangement of variables in a subsequence, and provides easy access to inherent semantic information embedded in the shape, characterized by the temporal order. 
Similar shapes generate point-pair distribution histograms having similar slope characteristics. 
Also, it must be observed that shape histogram is invariant to small shifts in onset of interesting state transitions. Small values of $m$,
and $n$ have been used in Figure \ref{fig11} for explanatory purposes, in practice higher values are used that give better results.

The pattern of time-series points exhibit strong correlation. Though small amount of stochastic element is present which makes future values
only partially related to past values. Nevertheless, for the subsequences characterizing a particular event or motif, the relative arrangement
of histogram points is unique. 

\begin{figure}[!h]
\centering
\subfloat[Subsequence1 from figure \ref{fig111} and its shape histogram computed with $m$ = 40, and $n$ =
36]{\label{seq1}\includegraphics[width
= 55mm]{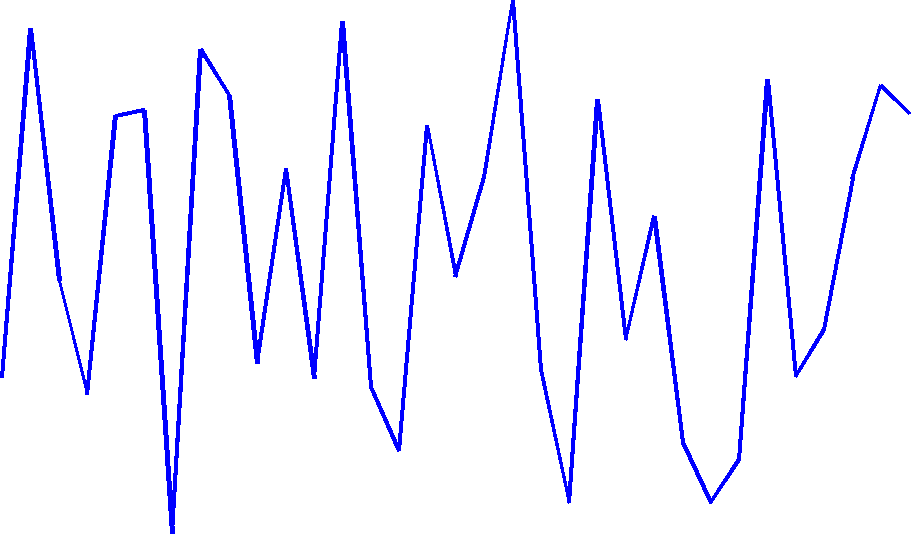} \includegraphics[width = 60mm]{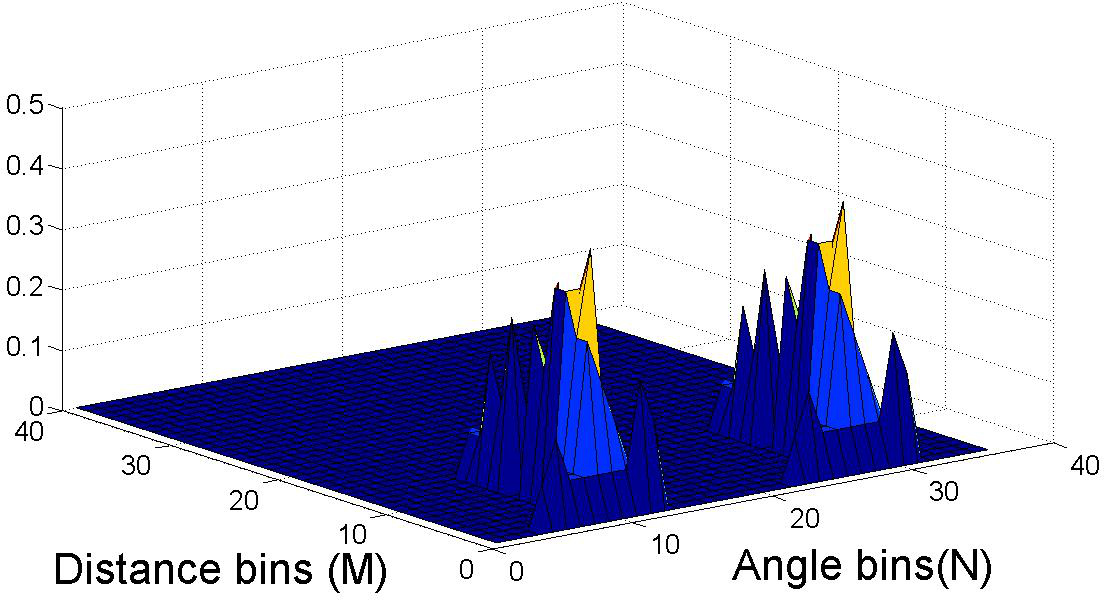}} \\
\subfloat[Subsequence2 from figure \ref{fig111} and its shape histogram computed with $m$ = 40, and $n$ =
36]{\label{seq2}\includegraphics[width
= 55mm]{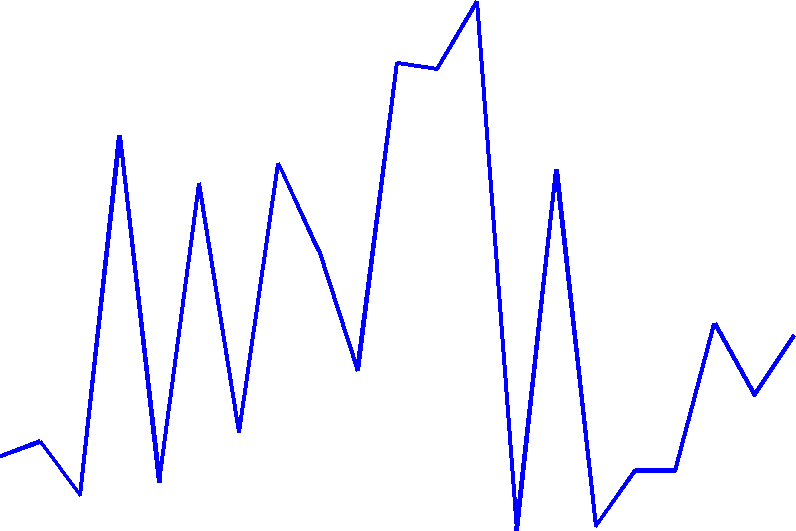}\label{seq2sd}\includegraphics[width = 60mm]{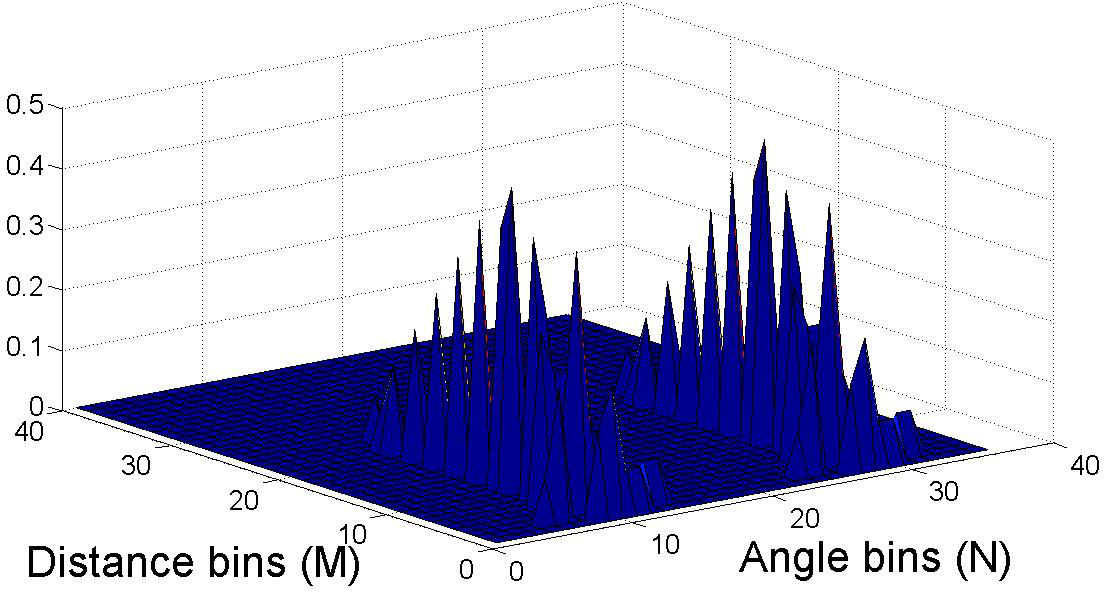}}
\caption{Two subsequences described in Figure \ref{fig111} and corresponding shape histogram (Subsequence 1 and Subsequence 2 shown in Figure \ref{fig111}) and corresponding shape histogram \label{fig31}}
\end{figure}

The global shape of a subsequence described by distantly positioned point-pairs is identified by the overall spread of the surface of $H$. The local shape of
a subsequence is described by peaks in $H$, as can be observed in Figure \ref{fig11}.

Parameters ($m$, $n$) control the robustness and discriminative ability of the shape histogram. In the supervised learning setting, parameter search by splitting the complete training data in training and validation set is the standard approach. 
However, this method is computationally complex, and requires sufficient labeled examples. Instead, we have evaluated the shape histogram
for range of distance and orientation quantization levels, and the best results are presented. 
Initial selection of ($m$, $n$) is based on the heuristic that a lower number of bins do not capture the uniqueness of point-pair distribution. Therefore, we select sufficiently large number of angular and distance bins such that the descriptor accurately captures the discriminative pattern of the point distribution histograms. 
The experimental evaluation showed that with increase in shape histogram parameters, the discriminative ability of feature increases with
increased similarity matching cost. However, for very large number of bins $(m, n)$, sparsely distributed point distribution histograms
are noisy and highly sensitive intra class variations. This is illustrated in Figure \ref{fig31} depicting a shape histogram using more
realistic values of parameters.

\subsection{Time and space complexity}
In a subsequence $S$ having $l$ values, we have $^lC_2$ unique-pairs requiring equal number of distance and angle computations for shape
histogram computation which are easy to compute. Shape histogram is of the dimension $m \times n$ requiring proportional storage
requirement, which is typically a small number. Therefore, while our method of using shape histogram has time complexity quadratic in the length
of the candidate subsequences, these are themselves of bounded length as compared to the length of the series itself. Thus our approach
is linear in the length of the overall time series. 

Further, the shape histogram is a low-dimensional feature vector that can be classified efficiently by standard classifiers.
Additionally, the histogram often has significant number of empty bins by the example shown in Figure
\ref{fig31} and substantiated by experimental results in the next Section. This property also
presents opportunities for further optimization from the efficiency perspective.

\section{Experimental Results}
\label{sec:exp}
To substantiate the key contributions of our work, we present the results from three different kinds of experiments, performed
on two types of datasets. 
First one was performed on a real-life dataset, obtained from the sensors of vehicles. 
This dataset results in variable length subsequences, with intermittently missing values. 
We present the results of this analysis in Section \ref{sec:e1}. 
Later, in Section \ref{sec:e2}, we present a summary of results indicating that it is possible to concatenate shape histogram of other sensors of the same electro-mechanical system, and that accuracy of event detection can be improved through this.
Thereafter, in Section \ref{sec:e3}, we present a summary and analysis of results obtained from the experiments performed on 24 different public datasets. Through these last set of experiments we demonstrate that our approach of using shape-histograms and classification mechanism such as SVM with RBF kernel, gives better results than many other approaches published in the research literature.

\subsection{Evaluation on vehicular sensor data}
\label{sec:e1}

We used our proposed shape histogram technique to detect events of interest from our vehicular sensor data as introduced in Section
\ref{sec:intro}. Our data collection consisted of measurements from seventeen different sensors. 
\textit{Event1} is characterized by dynamic behaviour of different sensors, nevertheless, we begin with  
a single sensor results using \textit{Sensor2} (speed) so as to illustrate the behaviour of our technique. Subsequently, we present results using other sensors as well,
as well as results using multiple sensors together, which significantly improves performance. 

\begin{table}[!h]
\centering
\caption{\textit{Event1} identification using \textit{Sensor2} }
\label{Tb_CarData}
\begin{tabular}{|c|c|c|}
\hline
\multicolumn{3}{|c|}{shape histogram}\\
\hline
Accuracy & Precision & Recall \\
\hline
96.52 & 94.41 & 88.05 \\
\hline
\multicolumn{3}{|c|}{shapelet$^{\ref{shapeletcode}}$}\\
\hline
89.3 & 81.04 & 85.18 \\
\hline
\end{tabular}
\end{table}

First 831 candidate subsequences potentially containing occurrences of \textit{Event1} were extracted from each sensor using the
rule-based approach as described in Appendix-A. These rules were relaxed in order to get 100\% recall, and the rules were drawn
based on domain knowledge. 
Here, by relaxed rules we mean, adjusting the threshold values in the rule, e.g., if one of the rules that characterizes the event is stated as, speed should be more than about 30 miles per hour, we could use 20 miles per hour to ensure 100\% recall.
The 831 subsequences varied in length from 9 to 39 time-steps. From this candidate set, 122 subsequences were manually tagged as containing occurrences of \textit{Event1} and 709 subsequences as false candidates, i.e., these did not contain a genuine occurrence of  \textit{Event1}. 

We first present results of event detection for \textit{Event1} using a single sensor, \textit{Sensor2}. This and all further experiments have been performed via 10-fold cross-validation using the manually tagged data as described above. 
A range of histogram parameters were tried out; in practice 15 distance bins and 40 orientation bins achieved the best results. Initial evaluation with a nearest neighbour classifier using the euclidean distance metric between shape histograms, or even a linear SVM classifier did not improve over results obtained via the rule based approach. 
However, using a radial (RBF) kernel with the SVM classifier showed significant improvement in the classification accuracy. We therefore, use Shape-Histograms + SVM with RBF kernel (ShapeHist-SVM-RBF) to benchmark the results of our approach with other approaches.

In order to compare our approach with other approaches, we have also applied shapelets\footnote{\label{shapeletcode}. The shapelet code used for our experiments was taken from \texttt{http://www.cs.ucr.edu/~mueen/LogicalShapelet/}}. \cite{Ye:2009} for classification of subsequences, for the same experimental setting. Table \ref{Tb_CarData} shows results of both the methods. As is evident, our approach of shape histograms coupled with SVM classifier using RBF kernel(ShapeHist-SVM-RBF) performs significantly better, at least on this particular time-series and event. We analyze the reasons for this improvement below.

\begin{figure}[!h]
\centering
\includegraphics[width=0.75\columnwidth]{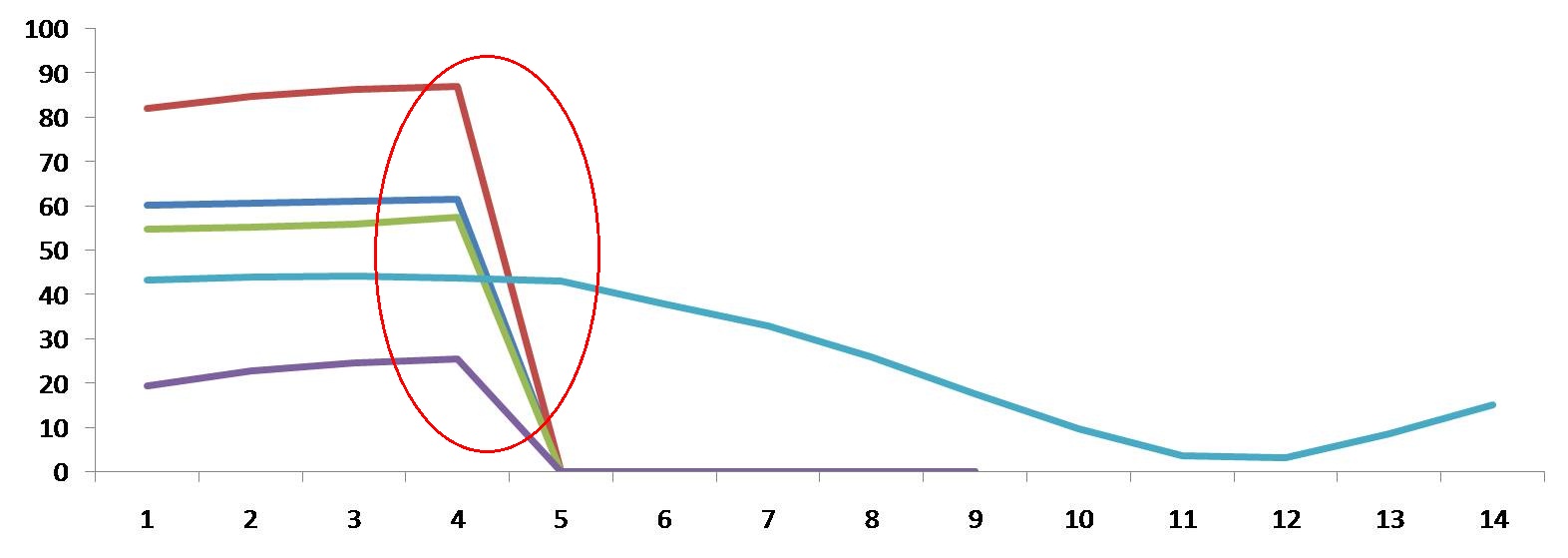} 
\caption{Wrongly classified examples of \textit{Event1}, by Shapelets, on subsequences of \textit{Sensor2}, shown by a red oval}
\label{Res_Car}
\end{figure}

Figure \ref{Res_Car} shows examples of \textit{Event1} occurrences missed by shapelet and correctly identified by shape histogram.
In general, we observed that shapelet failed in identifying the events generated by extreme braking conditions having low true positives and high true negatives (positives are the genuine \textit{Event1} events). 
An actual \textit{Event1} is generally characterized by high \textit{Sensor1}, and fast decrease in \textit{Sensor2} occurring for small time-duration. 
However, exceptional situations of prolonged engine acceleration may increase the duration of \textit{Event1} as shown in Figure \ref{Res_Car}. 

The shapelet based approach concentrates on a small local region, i.e., the region depicting the transition (shown by a red ellipse) for classification of subsequences. However, large intra-class variation does not provide sufficient distinction between categories in the decision tree architecture used by shapelet. 
Also, the decision about the occurrence of an event also requires the information of initial and final state of the sensor variable. 
In these cases, our shape histogram exploits the combination of local information, i.e., transition stage and global information i.e., initial and final state. 

We also observed some cases where the shape histogram approach wrongly classified the events, most often they were characterized by
relatively smooth decrease in \textit{Sensor2} measurement, i.e., a false \textit{Event1}. 
In general, we desire high recall in such problems because of subtle difference in patterns of real and false \textit{Event1}. 

\subsection{Linear Concatenation of Features}
\label{sec:e2}
As mentioned earlier and illustrated in Appendix-A, occurrences of \textit{Event1} actually involve the dynamics of multiple sensors. 
This requirement in real-life dataset and the fact that shape histograms of multiple sensors can be combined, 
make the use of shape histogram for detection of events in vehicular sensor data even more appropriate.
Therefore, we also evaluated the shape histogram approach for identifying \textit{Event1} from other sensors including \textit{Sensor4}, \textit{Sensor5}, and \textit{Sensor6}. 
These sensors represent acceleration, brake pressure, and deceleration measurements in the vehicle dataset. 
The results presented in Table \ref{Tb_SenRes} used experimentally obtained optimal shape histogram parameters for each of these sensors as shown in Table \ref{Tb_SenRes}.  

\begin{table}
\centering
\caption{\textit{Event1} detection: multiple sensors }
\label{Tb_SenRes}
\begin{tabular}{|p{1.1cm}|c|c|c|c|c|}
\hline
Sensor & \multicolumn{1}{p{1.2cm}|}{Accuracy} & \multicolumn{1}{p{1.2cm}|}{Precision} & \multicolumn{1}{p{1cm}|}{Recall} &
\multicolumn{1}{p{1cm}|}{Dist. bins} &\multicolumn{1}{p{1cm}|}{Ori. bins}\\
\hline
\textit{Sensor3} & 92.78 & 92.10 & 92.80 & 15 & 90 \\
\hline
\textit{Sensor4} & 95.05 & 94.80 & 95.10 & 25 & 20 \\
\hline
\textit{Sensor5} & 91.31 & 92.10 & 91.30 & 30 & 25 \\
\hline
\textit{Sensor6} & 92.25 & 91.40 & 92.20 & 30 & 20\\
\hline
\textit{Multi-Sensor} & 97.19 & 97.10 & 97.20 & - & - \\
\hline
\end{tabular}
\end{table}

Different sensors respond for a specific event with different dynamics, nevertheless, complete characterization of the event
requires information of complex interaction between responses of different sensor. 
Therefore we subsequently combined the shape histograms for all the sensors by linear concatenation. As also shown in Table \ref{Tb_SenRes}, identification of \textit{Event1} by combining multiple shape histograms for all these sensors showed 0.67\% gain in the identification accuracy. 
These results demonstrate that shape histograms can exploit information from multiple sensors to improve classification accuracy.

\begin{table*}
\scriptsize
\centering
\caption{Testing Accuracy taken from \cite{Lines:2012}, ShapeHist-SVM-RBF included with new ranks.}
\label{Tb_Rank}
\begin{tabular}{|p{1.8cm}|lr|lr|lr|lr|lr|lr|lr|lr|lr|}
\hline
Dataset Name & \multicolumn{2}{|p{0.2cm}|}{ShapeHist} & \multicolumn{2}{|p{1cm}|}{SVM Linear} &
\multicolumn{2}{|p{1cm}|}{Rotation Forest} & \multicolumn{2}{|p{1cm}|}{Random Forest} & \multicolumn{2}{|p{1cm}|}{Bayesian Network}
& \multicolumn{2}{|p{1cm}|}{Naive Bayes} & \multicolumn{2}{|p{1cm}|}{1-NN} & \multicolumn{2}{|p{1cm}|}{C4.5} &
\multicolumn{2}{|p{1cm}|}{Shapelet Tree}\\ 
\hline
Adiac & 28.25 & 4.0 & 23.79 & 9.0 & 30.69 & 1.0 & 30.43 & 2.0 & 25.06 & 7.0 & 28.13 & 5.0 & 25.32 & 6.0 & 24.3 & 8.0 & 29.92 & 3.0 \\
\hline
Beef & 73.33 & 4.5 & 86.67 & 2.0 & 70.00 & 6.0 & 60.00 & 7.5 & 90.00 & 1.0 & 73.33 & 4.5 & 83.33 & 3.0 & 60.00 & 7.5 & 50.00 & 9.0 \\
\hline
Chlorine-Concen. & 64.18 & 1.0 & 56.15 & 8.0 & 63.52 & 2.0 & 57.58 & 4.0 & 57.08 & 5.0 & 45.96 & 9.0 & 56.93 & 6.0 & 56.48 & 7.0 & 58.8 &
3.0 \\
\hline
Coffee & 100 & 2.5 & 100 & 2.5 & 89.29 & 8 & 100 & 2.5 & 96.43 & 5.5 & 92.86 & 7.0 & 100 & 2.5 & 85.71 & 9.0 & 96.43 & 5.5 \\
\hline
DiatomSize-Red. & 91.88 & 3.0 & 92.16 & 2.0 & 83.01 & 5.0 & 80.39 & 6.0 & 90.20 & 4.0 & 78.76 & 7.0 & 93.46 & 1.0 & 75.16 & 8.0 & 72.22 &
9.0 \\
\hline
ECGFiveDays & 99.60 & 1.0 & 98.95 & 3.0 & 98.61 & 4.0 & 93.26 & 8.0 & 99.54 & 2.0 & 96.4 & 6.0 & 98.37 & 5.0 & 96.17 & 7.0 & 77.47 & 9.0 \\
\hline
Face (four) & 96.73 & 6.0 & 97.73 & 4.5 & 98.86 & 3.0 & 87.5 & 7.0 & 100 & 1.5 & 97.73 & 4.5 & 100 & 1.5 & 76.14 & 9.0 & 84.09 & 8.0 \\
\hline
GunPoint & 98.67 & 3.5 & 100 & 1.0 & 98.67 & 3.5 & 96.00 & 6.0 & 99.33 & 2.0 & 92.00 & 7.0 & 98.00 & 5.0 & 90.67 & 8.0 & 89.33 & 9.0 \\
\hline
ItalyPower-Demand & 94.28 & 1.0 & 92.13 & 5.5 & 92.03 & 7 & 93.00 & 2.0 & 92.42 & 4.0 & 92.52 & 3.0 & 92.13 & 5.5 & 90.96 & 8.0 & 89.21 &
9.0 \\
\hline
Lighting7 & 64.38 & 4.5 & 69.86 & 1.0 & 65.75 & 2.5 & 64.38 & 4.5 & 65.75 & 2.5 & 57.53 & 6.0 & 49.32 & 8.5 & 53.42 & 7.0 & 49.32 & 8.5 \\
\hline
Medical-Images & 58.98 & 1.0 & 52.5 & 2.0 & 51.45 & 3.0 & 50.79 & 4.0 & 28.16 & 8.0 & 17.37 & 9.0 & 45.66 & 6.0 & 44.87 & 7.0 & 48.82 & 5.0
\\
\hline
Mote-Strain & 82.23 & 9.0 & 88.66 & 4.0 & 86.98 & 5.0 & 84.58 & 6.0 & 89.06 & 2.0 & 88.82 & 3.0 & 90.34 & 1.0 & 84.42 & 7.0 & 82.51 & 8.0 \\
\hline
Sony-AIBORob& 91.54 & 1.0 & 86.69 & 4.0 & 89.02 & 3.0 & 85.19 & 5.0 & 89.68 & 2.0 & 79.03 & 9.0 & 84.03 & 8.0 & 84.53 & 6.5 & 84.53 & 6.5 \\
\hline
Symbols & 64.97 & 8.0 & 84.62 & 4.0 & 84.42 & 4.0 & 84.62 & 4.0 & 92.26 & 1.0 & 77.99 & 6.5 & 85.63 & 2.0 & 47.14 & 9.0 & 77.99 & 6.5 \\
\hline
Synthetic-Cont. & 77.33 & 8.0 & 87.33 & 6.0 & 92.00 & 3.0 & 89.00 & 5.0 & 76.67 & 9.0 & 78.00 & 7.0 & 93.00 & 2.0 & 90.33 & 4.0 & 94.33 &
1.0 \\
\hline
Trace & 100 & 1.5 & 98.00 & 6.0 & 98.00 & 6.0 & 98.00 & 6.0 & 100 & 1.5 & 98.00 & 6.0 & 98.00 & 6.0 & 98.00 & 6.0 & 98.00 & 6.0 \\
\hline
TwoLead-ECG & 99.56 & 1.0 & 99.30 & 3.0 & 97.98 & 6.0 & 96.14 & 7.0 & 98.77 & 5.0 & 99.12 & 4.0 & 99.47 & 2.0 & 85.25 & 8.0 & 85.07 & 9.0 \\
\hline
\hline
Average Rank & \multicolumn{2}{|p{1cm}|}{\textbf{3.56}} & \multicolumn{2}{|p{1cm}|}{3.97} & \multicolumn{2}{|p{1.3cm}|}{4.24} &
\multicolumn{2}{|p{1cm}|}{5.09} & \multicolumn{2}{|p{1cm}|}{3.71} & \multicolumn{2}{|p{1cm}|}{6.09} &
\multicolumn{2}{|p{1cm}|}{4.18} & \multicolumn{2}{|p{1cm}|}{7.41} & \multicolumn{2}{|p{1cm}|}{6.76}\\
\hline
\end{tabular}
\end{table*}

\subsection{Evaluation on public datasets}  
\label{sec:e3}
For further evaluation our ShapeHist-SVM-RBF approach, we collected multiple datasets available in public domain. 
We took all of the available\footnote{\label{fn_na} Out of many datasets used in \cite{Lines:2012}, only 17 were available.} datasets from
\cite{Lines:2012}, and 7 additional datasets from \cite{dataset:Keogg}.  We present accuracy, precision, and recall for our approach on
these datasets in Table \ref{Tb_GunPoint} (Appendix-B). Number of Distance bins ($m$), Orientation bins ($n$), and average occupancy has
also been reported in the
same table.

It may appear that with the above parameter settings that the shape histogram for each subsequence is a vector having 1500 (25 $\times$ 60) real values. 
However, in practice on average only 4.08\% of total bins (Occupancy) across the instances have non-zeros entities which is of the order actual subsequence length; as a result storing and computing distances between shape histogram feature descriptors can be done efficiently.

To arrive at the results presented in Table \ref{Tb_GunPoint}, we experimented with a range of quantization levels for distances and orientations. 
First, a euclidean-distance-based nearest neighbor classifier as well as a linear kernel based support vector machine (SVM) was applied. 
In this case, shape histograms with 25 distance bins and 60 angle bins achieved better results in comparison with other parameter settings. 
However, using the L2-norm based similarity metric and SVM based classification using a linear decision boundary were less effective. 
Similar to how it worked on the real-life dataset, the best classification results were observed 
using SVM with the Gaussian kernel with standard deviation = 3 and cost parameter C = 3.33. 
The results corresponding these parameters are shown in Table \ref{Tb_GunPoint}.

Further, to compare the results of our approach with other approaches presented in the research literature, we took the table (Table 3 of \cite{Lines:2012}) that reports the accuracy of various approaches in \cite{Lines:2012} on multiple datasets. 
This is because, to the best of our knowledge work done by Lines et al. in \cite{Lines:2012} by and large reports the best accuracies of
various approaches of event detection on time-series. 
To execute this comparison we took only the 17 datasets$^{\ref{fn_na}}$ that were common between our and their experiments, included our
results (ShapeHist-SVM-RBF) in the table, and recalculated the ranks as shown in Table \ref{Tb_Rank}. For calculation of these ranks we
assigned `average ranks' to the algorithms that perform equally well, e.g., in Coffee dataset, 4 approaches give 100\% accuracy, we
assign them all a rank of $(1+2+3+4)/4$. Unlike \cite{Lines:2012}, our method of assigning these ranks is same as
how ranks are assigned in Wilcoxon signed rank test\cite{p:wilcoxonTest}. We also
performed the Wilcoxon signed rank test and were able to reject the null hypothesis for Naive-Bayes, Shapelets, and C4.5. 
However, when we performed z-test on the two most competing approaches (ShapeHist and Bayesian-Network), we were able to reject the null
hypothesis indicating that the mean of the two approaches are not same.

\begin{figure}[!h]
\centering
\includegraphics[width=0.65\columnwidth]{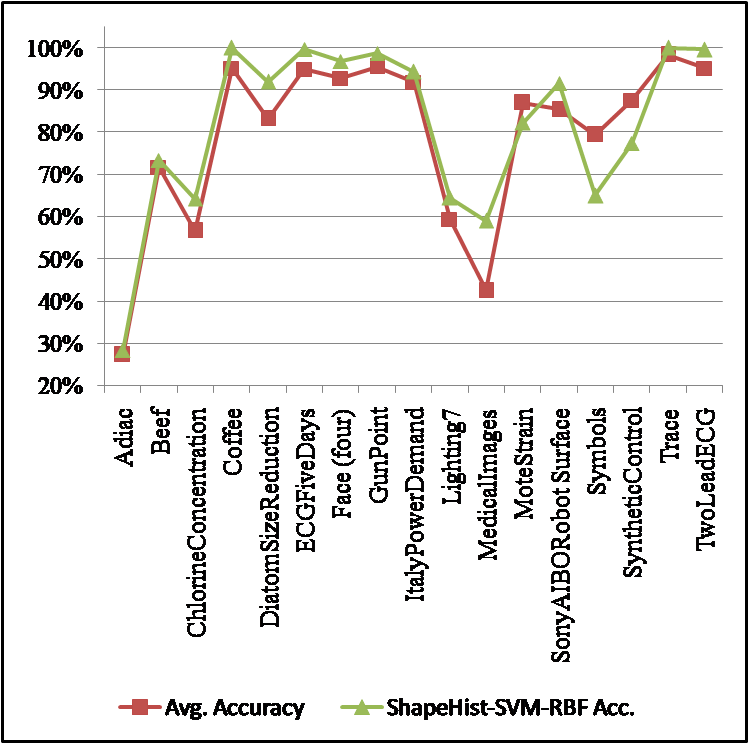}
\caption{Average accuracy of various approaches reported in \cite{Lines:2012}, compared with the accuracy of
ShapeHist-SVM-RBF; on 17 common$^{\ref{fn_na}}$ datasets.}
\label{fig_avg}
\end{figure}

\begin{figure}[!h]
\centering
\includegraphics[width=0.65\columnwidth]{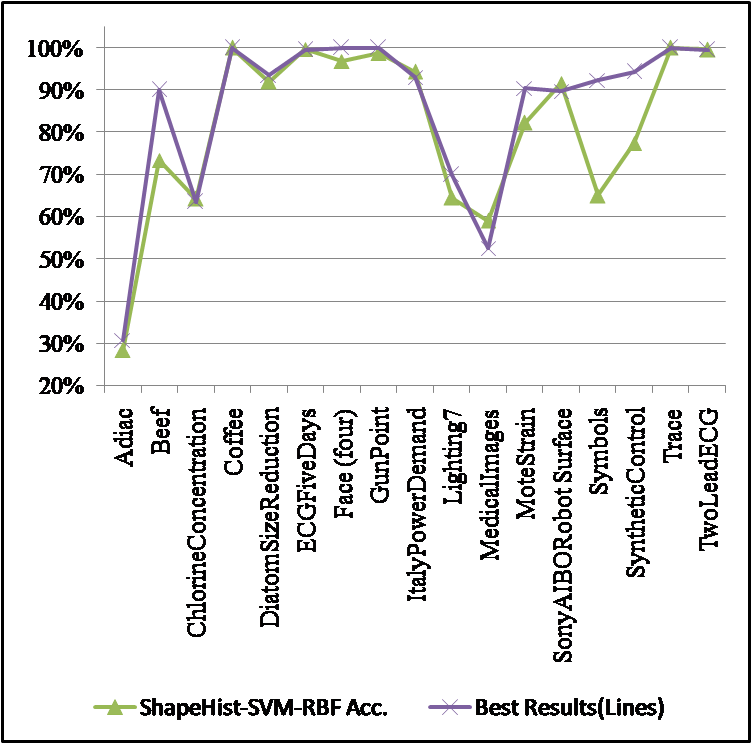}
\caption{The best accuracy of various approaches reported in \cite{Lines:2012}, compared with the accuracy of
ShapeHist-SVM-RBF; on 17 common$^{\ref{fn_na}}$ datasets.}
\label{fig_bst}
\end{figure}

We plotted a comparison of our approach with the average accuracy of various approaches, on the 17 common$^{\ref{fn_na}}$ datasets as shown
in Figure \ref{fig_avg}(Appendix-B). It can be observed from this plot that our approach performs consistently better than other approaches
barring a few cases. Similarly, we 
also plotted a comparison of our approach with the best accuracy of any approach reported in \cite{Lines:2012}, on the 17
common$^{\ref{fn_na}}$ datasets also shown in Figure \ref{fig_bst}. From this, we observe that our approach is almost similar to the best
performing approach on multiple datasets. (Note: Single approach that performs the best on all these datasets does not exist.)
Also, on other datasets such as SwedishLeaf and Lighting2 that are not included in \cite{Lines:2012}, our approach performs better.
Further, from these observations and from Table \ref{Tb_Rank}, it can be inferred that by and large \textbf{the
average rank of our approach is 3.56, which is the best, making it the most versatile approach for detection of events from time-series
data}.
\section{Related work}
\label{sec:rw}
Detecting events of interest in time-series data has been studied in many different application domains. Initial work explored a variety of
methods and classifiers based on statistics and machine learning theory. Guralnik and Srivastava \cite{Guralnik:1999} addressed the time
series event detection as a change detection problem by modeling the time-series using a piecewise segmented model. However, they
focus on change point detection and divide the whole time-series in different segments based on these change points. Our problem is different
in that our events of interest are more complex than change points alone, e.g., they may involve steady regions as well as sharp dips, etc.

Early work by Rodriguez et al. \cite{rodriguez00learning} introduced inductive logic programming for multivariate time-series classification
as well as the application of boosting for performance improvement of the classifiers. Their approach worked with variable
length event patterns however it required expert intervention for definition of a region. A knowledge-based event detection
framework for time-series arising in health monitoring systems is discussed in \cite{Hunter99knowledge}. Other approaches include using
recurrent neural networks for time-series classification by organizing the network dynamics in alignment with class labels\cite{HuskenS03}.
Availability of sufficient labeled data is important aspect of time-series classification problem. In this
context, Wei and Keogh \cite{Wei:2006} presented semi-supervised learning using nearest neighbor time-series classification with
insufficient labeled data. However, all these approaches work with fixed length subsequences only.

Xi et al. \cite{Xi:2006} proposed a dynamic-time-warping-based nearest neighbor classifier for addressing time complexity in case of large
datasets. Recently Jeong et al. \cite{Jeong2011} presented weighted dynamic time warping for time-series classification by associating
higher weights to phase differences in local neighborhood. Such approach suffer from issues of efficiency and it would be hard to process
large time-series datasets. In summary, each of the above methods concentrated on experimentation with different classifiers and similarity
measures. However, they did not adequately address real-time requirements of time and space complexity. 

In \cite{Ye:2009}, Ye and Keogh introduced `shapelet' based time-series classification, where a shapelet is a time-series sub-sequence
considered as maximal representative of a class. Information-gain based shapelet discovery locates the most informative local pattern for
classification using linear search over the complete time-series. Subsequently a decision tree is defined on the computed information gains
to perform the time-series classification. This approach suffers from high time-complexity.

In an extension of shapelet based time-series analysis, Mueen et al. \cite{Mueen:2011} proposed logical combination of multiple shapelets
for performance improvement. Lines et al. \cite{Lines:2012} extended the shapelet concept for transforming the time-series in uniform data
space so as to be able to use other classifiers in addition to decision trees. The \textit{k} best shapelets extracted from the data in
combination with decision tree classifier were shown to achieve comparable performance with the results presented in \cite{Ye:2009}. In
further extensions of the shapelet idea of local pattern analysis, Zakaria et al. \cite{Zakaria2012} proposed unsupervised learning based
shapelet discovery method for clustering time-series subsequences. The underlying philosophy is the creation of min-dist based distance map
between a local pattern (read unsupervised shapelet), and a set of time-series subsequences.

Most of the above pattern detection techniques deal with detecting fixed-length waveform patterns. Shapelets, however, do treat
variable-length patterns as does our shape histogram approach; therefore our experimental comparisons in Section \ref{sec:exp} are primarily
with the shapelet approach, especially on our real-life vehicular data where variable length events are indeed present.
At the same time, since the shapelet descriptors are themselves of variable length this restricts their use in conjunction with state-of-the
art machine learning classifiers such as SVMs; in contrast, shape histograms do not have this restriction. (Note that this advantage that is
also shared by the shapelet-transform extension by Lines et al. \cite{Lines:2012}.)
Also, the shapelet discovery algorithms discussed in \cite{Ye:2009,Mueen:2011,Lines:2012} require apriori label information: in
other words, they require labelled data to calculate the shapelet descriptors themselves, rather than just for the classification stage. In
contrast our shape histogram feature descriptor can be computed independent of the events to be detected, i.e., un-supervised feature
extraction, and still performs better on real data.

\section{Conclusions}
We have proposed the shape histogram, a novel feature descriptor for waveform patterns inspired by ideas from image processing and computer vision, and shown that it is useful for event identification in time-series. 
The shape histogram is able to tackle scenarios involving varying duration events as well as is able to handle sparsity in data caused by missing values. 
We have presented experimental results on real-life vehicular sensor data and shown that our approach is superior to available implementations of previous pattern detection approaches. We have also evaluated our technique using publicly available data and shown that it achieves results comparable with the best previous
results on those datasets. Our method presents novel extension of shape based feature extraction techniques to time-series data representation
and demonstrates the applicability of shape-based features derived from image processing for event detection in real-life time-series data, especially that
arising in the context of vehicular sensor analysis.  
\bibliographystyle{abbrv}
\bibliography{bibtex/paper}

\section*{Appendix - A}
An interesting event in vehicular sensor data might be as described colloquially by an engineer as follows:
``\textit{Event 1} is characterized by a sudden high deceleration of a vehicle
running at high speed reaching an almost stationary position within 2-3 seconds after applying breaks.'' 
A hand-crafted rule for detecting the point of exact state transition, i.e., critical point during \textit{Event1} might
characterize such an event as follows:
 
 \begin{itemize}
\item[ ] \{(\textit{Sensor1} is greater than $s_1$) $\bigcap$ (\textit{Sensor2} more than $s_2$) $\bigcap$ (\textit{Sensor3} is in [$s_{31},
\ldots, s_{3m}$]  ) $\bigcap$ (drop in \textit{Sensor2} is more than $s_{2}^{*}$) \}
\end{itemize}

\textit{Sensor1}, \textit{Sensor2}, and \textit{Sensor3} refer values of brake cylinder pressure, wheel speed and gear as sampled in
real-time by onboard sensors. Subsequently, heuristic rules help us identify the left and right boundary of these critical points, to
identify subsequences that contain occurrence of \textit{Event1}. For example, we might define the extreme
boundaries of the event sequence in terms of the following conditions:
\begin{itemize}
\item[ ] Left boundary: \{(\textit{Sensor1} is less than $a$ for consecutive $b$ seconds) $\bigcup$ ($c$ seconds before the critical
point)\}
\item[ ] Right boundary: \{(\textit{Sensor1} is less than $d$ for consequent $e$ seconds) $\bigcap$ (\textit{Sensor2} in more than $f$ for
consecutive $g$ seconds) $\bigcup$ (35 Points before the critical point)\}
\end{itemize}
Figure \ref{Ex_events} depicts time series data of \textit{Sensor2} for two runs of a vehicle along with occurrences of possible
\textit{Event1} defined in rectangular windows. When such an approach based on hand-crafted rules were used for event detection, we found
that it gave poor results; e.g., we observed barely 60-70 \% accuracy
for \textit{Event1}. Further, different sets of rules are used by each engineer, which makes it difficult to arrive at principled conclusions
based on analysis of field data. 

However, if instead we use rules similar to those above but with more relaxed conditions (i.e., different, less conservative values of $s_1, s_2, s_3, a, b, c, d, e, f, g$),
we get can 100\% recall, but very poor precision: Exactly such relaxed rules
are used to generate \textit{candidate subsequences}, as referred to in Sections 2.1 and 3.2, on which our shape-histogram-based event
detection procedure is then applied.
\section*{Appendix - B}

\begin{table*}[!t]
\scriptsize
\centering
\caption{Time-series classification Shape histogram on standard datasets}
\label{Tb_GunPoint}
\begin{tabular}{|l|l|c|c|c|c|c|c|}
\hline
S.N. & Dataset & \multicolumn{1}{p{1.8cm}|}{Distance bins ($m$)} & \multicolumn{1}{p{1.8cm}|}{Orientation bins ($n$)} & Accuracy & Precision
& Recall & \multicolumn{1}{p{1.8cm}|}{Average Occupancy}\\
\hline
\hline
1. & Adiac \cite{Adiac_2005} & 25 & 36 & 28.25 & 28.44  & 28.18 & 5.63  \\
\hline
2. & Beef \cite{BagnallDHL12} & 45 & 10 & 73.33 & 74.64 & 73.68 & 13.10 \\
\hline
3. & ChlorineConcentration \cite{Li:2009} & 30 & 30 & 64.18 & 60.64 & 63.97 & 9.12 \\
\hline
4. & Coffee \cite{Lines:2012} & 35 & 45 & 100 & 100 & 100 & 7.27 \\
\hline
5. & DiatomSizeReduction \cite{dataset:Keogg} & 32 & 40 & 91.88 & 92.34 & 91.12 & 6.79 \\
\hline
6. & ECG \cite{Olszewski:2001,Ye:2009,Lines:2012} & 40 & 45 & 91.35 & 90.64 &  90.87 & 3.76 \\
\hline
7. & ECGFiveDays \cite{Olszewski:2001,Ye:2009,Lines:2012} & 27 & 40 & 99.60 & 99.37 & 99.48 & 6.14 \\
\hline
8. & Face (four) \cite{Face_dataset} & 36 & 90 & 96.73 & 95.83 & 97.12 & 5.16 \\
\hline
9. & FacesUCR \cite{dataset:Keogg} & 25 & 60 & 86.56 & 86.44 & 85.66 & 4.14 \\
\hline
10. & GunPoint \cite{Ye:2009,RatanamahatanaK04} & 25 & 60 & 98.67 & 97.63 & 97.63  & 4.06\\
\hline
11. & ItalyPowerDemand \cite{Lines:2012} & 36 & 36 & 94.28 & 94.26 & 94.18 & 4.86 \\
\hline
12. & Lighting2 \cite{Lin:2007,4634188,libsvmDE02a} & 40 & 36 & 88.64 & 88.91 & 87.64 & 8.61 \\
\hline
13. & Lighting7 \cite{Lin:2007,4634188,libsvmDE02a} & 40 & 40 & 64.38 & 65.11 & 64.76 & 4.03 \\
\hline
14. & MedicalImages \cite{dataset:Keogg} & 35 & 36 & 58.98 & 56.46 & 58.10 & 6.38 \\
\hline
15. & MoteStrain \cite{Ye:2009,Lines:2012} &  25 & 45 & 82.23 & 82.73 & 82.08 & 5.37 \\
\hline
16. & OliveOil \cite{BagnallDHL12} & 40 & 36 & 80.00 & 79.84 & 80.13 & 10.48 \\
\hline
17. & SonyAIBORobot Surface \cite{dataset:Keogg} & 35 & 15 & 91.54 & 89.91 & 91.46 & 8.23 \\
\hline
18. & SonyAIBORobot SurfaceII \cite{dataset:Keogg} & 25 & 10 & 88.88 & 89.11 & 89.02 & 9.66 \\
\hline
19. & SwedishLeaf \cite{Xi:2006} & 20 & 60 & 74.53 & 74.78  & 74.24 & 6.41 \\ 
\hline
20. & SyntheticControl \cite{Alcock99} & 30 & 60 & 77.33 & 78.10 & 78.63 & 3.88 \\
\hline
21. & Symbols \cite{dataset:Keogg} & 35 & 36 & 64.97 & 64.76 & 64.39 & 7.94 \\
\hline
22. & Trace \cite{RatanamahatanaK04,Roverso00,dataset:Keogg} & 35 & 36 & 100 & 100 & 100 & 7.64 \\
\hline
23. & TwoLeadECG \cite{Olszewski:2001,Ye:2009,Lines:2012} & 36 & 36 & 99.56 & 99.45 & 99.58 & 4.45\\
\hline
24. & Wafer \cite{Wei:2006,dataset:Keogg} & 50 & 45 & 99.68 & 99.90 & 99.58 & 6.87 \\
\hline
\end{tabular}
\end{table*}

\end{document}